\pdfoutput=1
\documentclass[11pt]{article}

\usepackage[final]{acl}

\usepackage{times}
\usepackage{latexsym}
\usepackage{amsmath}
\usepackage{xcolor}
\usepackage{enumitem}
\usepackage[T1]{fontenc}
\usepackage[utf8]{inputenc}

\usepackage{microtype}
\usepackage{graphicx}
\usepackage{subcaption}

\usepackage[ruled, lined, linesnumbered, commentsnumbered, longend]{algorithm2e}

\definecolor{hotpink}{RGB}{255, 83, 115}
\definecolor{teal}{RGB}{90, 200, 250}
\definecolor{lightgreen}{RGB}{33, 222, 128}
\definecolor{lightblue}{RGB}{72, 123, 232}
\definecolor{blue}{RGB}{52, 52, 235}

\title{Cross-Modal Attribute Insertions for Assessing the Robustness of Vision-and-Language Learning}

\author{Shivaen Ramshetty$^*$ \and Gaurav Verma$^*$ \and Srijan Kumar 
\\
Georgia Institute of Technology
\\
\texttt{ \{sramshetty3, gverma, srijan\}@gatech.edu}
}

\begin{document}
\maketitle
\begingroup\def\thefootnote{*}\footnotetext{Equal contribution.}\endgroup
\begin{abstract} 
The robustness of multimodal deep learning models to realistic changes in the input text is critical for their applicability to important tasks such as text-to-image retrieval and cross-modal entailment.
To measure robustness, several existing approaches edit the text data, but do so without leveraging the
cross-modal information present in multimodal data.
Information from the visual modality, such as color, size, and shape, provide additional attributes that users can include in their inputs. 
Thus, we propose cross-modal attribute insertions as a realistic perturbation strategy for vision-and-language data that inserts visual attributes of the objects in the image into the corresponding text (e.g., ``girl on a chair'' $\rightarrow$ ``little girl on a wooden chair''). 
Our proposed approach for cross-modal attribute insertions is modular, controllable, and task-agnostic. We find that augmenting input text using cross-modal insertions causes state-of-the-art approaches for text-to-image retrieval and cross-modal entailment to perform poorly, resulting in relative drops of $\sim15\%$ in MRR and $\sim20\%$ in $F_1$ score, respectively. Crowd-sourced annotations demonstrate that cross-modal insertions lead to higher quality augmentations for multimodal data than augmentations using text-only data, and are equivalent in quality to original examples. We release the code to encourage robustness evaluations of deep vision-and-language models: \url{https://github.com/claws-lab/multimodal-robustness-xmai}. 
\end{abstract}

\section{Introduction}
The ability to model the interaction of information in vision and language modalities powers several web applications --- text-to-image search~\cite{he2016cross}, summarizing multimodal content~\cite{zhu2018msmo}, visual question answering~\cite{antol2015vqa}, and editing images using language commands ~\cite{shi2021learning}. 
Ensuring satisfactory user experience within such applications necessitates the development of multimodal models that can \textit{robustly} process text and image data, jointly.

\begin{figure}[!t]
    \centering
    \includegraphics[width=1.0\linewidth]{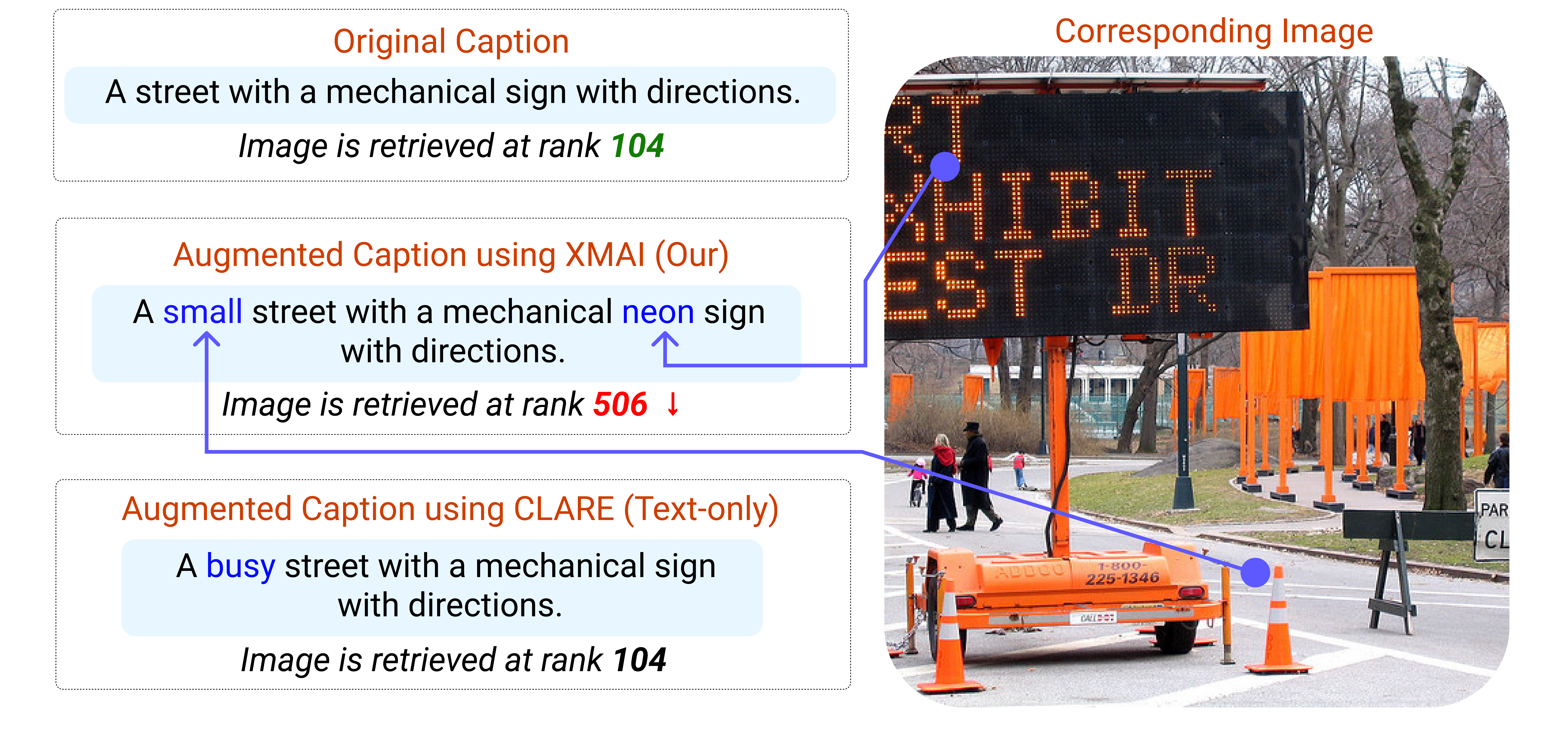}
    \caption{{We propose Cross-Modal Attribute Insertions (XMAI) --- an approach that leverages cross-modal interactions in multimodal data to obtain meaningful text augmentations that methods using text-only information (e.g., CLARE) cannot provide. These augmentations highlight vulnerabilities of multimodal models; in this case, the corresponding image is retrieved at a worse rank ($104$ $\rightarrow$ $506$) for the modified caption.}}
    \label{fig:overview}
\end{figure}

Existing research has demonstrated the brittle reasoning mechanism of text-only and image-only models by introducing variations in the inputs~\cite{evtimov2020adversarial, li-etal-2021-contextualized}. Furthermore, prior work have established controlled generation methods for text ~\cite{ross-etal-2022-tailor}, including counterfactuals for model assessment~\cite{Madaan_Padhi_Panwar_Saha_2021, wu-etal-2021-polyjuice}. However, beyond applying modality-specific perturbations to multimodal (image + text) data~\cite{qiu2022multimodal} , existing research has not studied the robustness of models to \textit{likely} augmentations in text that leverage cross-modal interactions.
Specifically, current research on text augmentation considers the following likely variations:  skipping certain words, introducing typographical errors, inserting noun or verb modifiers, or using synonyms. Consequently, to study the robustness of deep models, several automated methods have been developed to introduce these variations in the text. 
However, while these \textit{text-only} perturbations can cover more variations, they are by no means exhaustive with respect to multimodal data. In the context of multimodal data, the text accompanying an image can be meaningfully perturbed to include information from the image. For instance, users can issue a query on a search engine that specifies attributes of the desired image(s); `\textit{a male driver posing with a red car}' instead of `\textit{a driver posing with a car}.' Existing augmentation approaches can only model text-only data and cannot introduce relevant cross-modal information (like `male' and `red' in the above example) while generating augmentations.

We propose novel text variations that leverage the image modality to insert relevant information into the text, which we call \textit{cross-modal attribute insertions}. Our method inserts attributes of objects that are both present in the image and mentioned in the text. To do so, \textit{cross-modal attribute insertion} uses object detection to capture objects and their attributes in the image~\cite{anderson2018bottom}, and masked-language modeling to place those attributes prior to the object's mentions in the text~\cite{devlin-etal-2019-bert} (see Figure \ref{fig:overview}). Additionally, we use embedding similarities to expand the search space of possible augmentations, and introduce an adversarial component to estimate the robustness of multimodal models. 

Our proposed approach is highly modular, controllable, and task-agnostic. Different modules govern attribute selection from images, cross-modal object matching, attribute insertion in text, and adversarial strength of the augmented example. The contribution of these modules toward the final augmented text can be controlled using weights that can be tuned as hyper-parameters. Finally, our approach for generating augmentations does not involve any parameter training, which makes it task-agnostic and broadly applicable.

We demonstrate the applicability of our cross-modal attribute insertion approach by generating augmentations for assessing the robustness of models for two different multimodal tasks --- \textit{(a)} text-to-image retrieval and \textit{(b)} cross-modal entailment. Together, these two tasks are representative of ranking and classification multimodal tasks.
Our evaluation comprises assessing the robustness of state-of-the-art multimodal learning approaches for these tasks to our augmentations as well as quantifying the relevance of generated augmentations to unmodified examples. We contrast our cross-modal attribute insertions with several baseline approaches that model text-only information. 

\noindent Our key contributions and findings are:\\
    $\bullet$ We propose cross-modal attribute insertions as a new realistic variation in multimodal data. Our proposed approach introduces these variations in a modular, controllable, and task-agnostic manner. \\
    $\bullet$ We demonstrate that state-of-the-art approaches for text-to-image retrieval and cross-modal entailment are not robust to cross-modal attribute insertions, demonstrating relative drops of $\sim15\%$ and $\sim20\%$ in MRR and $F_1$ score, respectively.\\
    $\bullet$ While being as effective as existing text-only augmentation methods in highlighting model vulnerabilities, our approach produces augmentations that human annotators perceive to be of better quality than the most competitive text-only augmentation method. Furthermore, our method matches the quality of unmodified textual examples, while being at least $9\times$ faster than the most competitive baseline across the two multimodal tasks.
    
Overall, we find that cross-modal attribute insertions produce novel, realistic, and human-preferred text augmentations that are complementary to current text-only perturbations, and effectively highlight the vulnerabilities of multimodal models. Future work could employ our augmentation strategy to evaluate and develop more robust vision-and-language models.

\begin{figure*}
    \centering
    \includegraphics[width=1.0\linewidth]{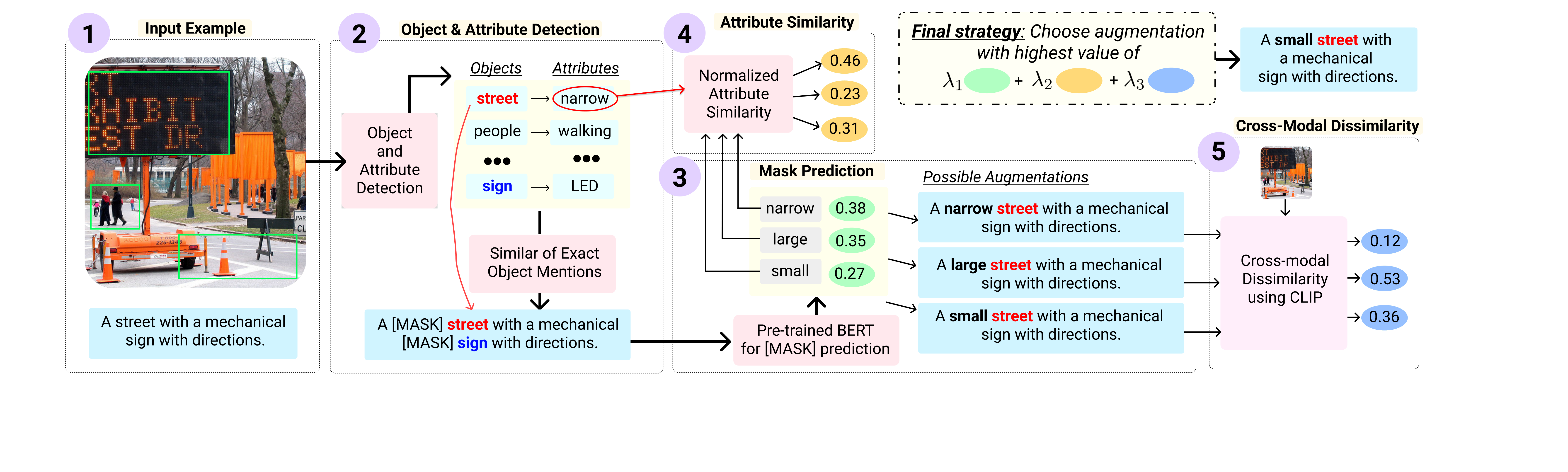}
    \caption{{\textbf{Schematic depicting cross-modal attributes insertion}. For each input example \textbf{\textit{(1)}}, we find objects that are depicted in the image and also mentioned in the text \textbf{\textit{(2)}}. Masked-language modeling is used to predict potential words that could describe common objects, leading to candidate augmentations \textbf{\textit{(3)}}. However, the final strategy takes into consideration the similarity of predicted words with the detected attribute for the object \textbf{\textit{(4)}} and the cross-modal dissimilarity between augmented text and the original image \textbf{\textit{(5)}} using $\lambda_i$ hyper-parameters. The augmentation strategy is also presented as an algorithm in Appendix  Alg. \ref{alg:two}.}}
    \label{fig:xmi_simple_overview}
\end{figure*}

\section{Related Work}

\noindent\textbf{Text Augmentations in Multimodal Data}:
Existing research investigating the robustness of deep learning models for natural language processing has proposed several automated approaches to introduce plausible variations in the text. 
~\citeauthor{ribeiro-etal-2020-beyond} (\citeyear{ribeiro-etal-2020-beyond}) and ~\citeauthor{naik18coling} (\citeyear{naik18coling}) propose a comprehensive list of perturbations that NLP models should be robust to --- including distracting phrases, URLs, word contractions and extensions. Many of these perturbations are task-agnostic and hence can be used to modify the text in multimodal (image + text) data as well. Similarly, other task-agnostic approaches to modify text data include random deletion, swapping, and insertion of words~\cite{wei-zou-2019-eda} and replacing, inserting, and merging words or phrases using masked language modeling~\cite{li-etal-2021-contextualized}. TextAttack~\cite{morris2020textattack} provides a comprehensive categorization of such methods and a framework to implement them. However, these methods lack in two critical ways: \textit{(i)} automated text augmentations often compromise the semantic meaning to notable
extents~\cite{wang2021adversarial}, and (ii) they only rely on the information contained in the text modality. In this work, we introduce augmentations in the textual part of multimodal data using TextAttack methods and consider them as \textit{baseline} augmentations. Then to overcome the flaws mentioned, we propose an approach that leverages the information in the visual modality to insert visual attributes in the textual modality (i.e., cross-modal attribute insertions). 

\noindent\textbf{Robustness of Multimodal Models}: Previous studies independently introduce unimodal perturbations in the visual or textual part of the input to study multimodal robustness. This could involve introducing imperceptible adversarial noise in the images and independently modifying the text using the augmentation approaches discussed earlier~\cite{li-etal-2021-contextualized, ribeiro-etal-2020-beyond, wei-zou-2019-eda}. For instance, ~\citeauthor{chen2020counterfactual} (\citeyear{chen2020counterfactual}) synthesize counterfactual samples of multimodal data using language models to modify the text. To ensure the preservation of the semantic meaning in the augmented text, ~\citeauthor{sheng2021human} (\citeyear{sheng2021human}) and ~\citeauthor{li2021adversarial} (\citeyear{li2021adversarial}) employ humans to perturb the textual questions to fool the state-of-the-art models for Visual Question Answering~\cite{antol2015vqa}. In a step towards using cross-modal interactions in image-text data to generate realistic variations,~\citeauthor{verma-etal-2022-robustness} (\citeyear{verma-etal-2022-robustness}) proposed adding relevant information from the image to expand the original textual description, and assess the robustness of multimodal classifiers.  Our work proposes a different approach to leverage cross-modal associations to augment multimodal data. Instead of expanding the original text, we \textit{insert attributes} of objects in the image that are also mentioned in the corresponding text to modify the original text. Additionally, our work considers more multimodal tasks by studying text-to-image retrieval and cross-modal entailment.

\section{Cross-Modal Attribute Insertions}
\label{sec:XMAI}
Our approach for augmenting text in multimodal data involves identifying objects in the image that are also mentioned in the text, and inserting words similar to their attributes in the text at relevant places. An overview of our approach (XMAI) is depicted in Figure \ref{fig:xmi_simple_overview}. 

We denote paired multimodal units as $(\mathcal{I}, \mathcal{T})$, where $\mathcal{I}$ represents the input image and $\mathcal{T}$ is the text corresponding to that image. Our goal is to transform $\mathcal{T}$ into $\mathcal{T}'$ such that the text includes relevant information from $\mathcal{I}$ while effectively highlighting the target model's vulnerabilities. 
Our method to infuse object attributes in the text  can be broken into four parts: \textit{(a)} object and attribute detection in $\mathcal{I}$, \textit{(b)} BERT-based \texttt{[MASK]} prediction in $\mathcal{T}$ while ensuring \textit{(c)} similarity of inserted tokens with detected object attributes, and \textit{(d)} enforcing dissimilarity between modified text $\mathcal{T}'$ and $\mathcal{I}$ to obtain robustness estimates of multimodal models.

\vspace{0.05in}
\noindent\textbf{Object and Attribute Detection}: For each image $\mathcal{I}$ we use a pre-trained bottom-up attention model \cite{anderson2018bottom, yu2020buapt} to extract objects and their associated attributes. The bottom-up attention model identifies objects and corresponding attributes with a one-to-one mapping.  We use these objects and attributes to modify $\mathcal{T}$, by introducing masks (i.e., the \texttt{[MASK]} token), in front of the mentions of the objects in $\mathcal{T}$. However, a strict matching criterion would ignore similar objects or alternatively named objects in $\mathcal{T}$. To address this, whenever the text does not have any direct object matches we use a Parts of Speech (PoS) tagger to identify nouns that could represent image objects. 

These identified nouns are compared to objects using cosine similarity between the word embeddings. If the cosine similarity between a noun $\mathcal{T}$ and a detected object in $\mathcal{I}$ is above some threshold, $t$, then a \texttt{[MASK]} token is placed before that noun. Overall, this step ensures that insertions are made only for objects in $\mathcal{T}$ that are seen in the corresponding image $\mathcal{I}$ to obtain $\mathcal{T}'$.

\vspace{0.05in}
\noindent\textbf{Mask Prediction}: Next, we aim to fill in the \texttt{[MASK]} tokens with contextually relevant object attributes. To do so, we use the pre-trained language model BERT \cite{devlin-etal-2019-bert}. We sample top-$k$ predictions from the BERT model  based on probability scores that also meet the following criteria: the predicted word should not be a stop word and should not exist in the 3-hop neighborhood of the current \texttt{[MASK]}. Furthermore, since $\mathcal{T}$ may contain more than one \texttt{[MASK]} token, we carry out this process sequentially for each \texttt{[MASK]} to utilize newly introduced contexts. Following this process, we obtain $k$ candidate insertions that are contextually relevant for each of the identified objects in $\mathcal{T}$ that also exists in $\mathcal{I}$. In the next step, to maintain cross-modal relevance, we consider the similarity of these candidate attributes with the attributes detected in $\mathcal{I}$.

\vspace{0.05in}
\noindent\textbf{Attribute Similarity}:
To better select a word for a specific mask that aligns well with information in $\mathcal{I}$, we only consider predicted tokens similar to the attributes of the associated object detected in $\mathcal{I}$. In order to do so, we utilize embedding-based similarities between each predicted token and the detected attribute string. The image attributes can describe the shape, size, color, or other characteristics (like `\textit{floral} dress') of detected objects.

\vspace{0.05in}
\noindent\textbf{Cross-Modal Dissimilarity for Estimating Robustness}:
Finally, to estimate the robustness of multimodal models, we explicitly include a component that encourages dissimilarity in the embeddings of the image $\mathcal{I}$ and the modified text $\mathcal{T}'$. For each possible modified text $\mathcal{T}'$, we compute the cosine distance between its embedding obtained using the CLIP model~\cite{radford2021learning} and that of the corresponding image's CLIP embedding. While the mask prediction and attribute similarity steps ensure that the attribute insertions are semantically meaningful and maintain cross-modal relevance, the cross-modal dissimilarity between $\mathcal{T}'$ and $\mathcal{I}$ ensures that we leverage the vulnerabilities in the encoding mechanism of multimodal models. We use CLIP as the encoder for this step as it is a strong representative of the state-of-the-art vision-language models. 

\vspace{0.05in}
\noindent\textbf{Text Augmentation Strategy}: 
We now choose the final augmentation of $\mathcal{T}$ by combining the above four components ––– object and attribute detection, mask prediction, attribute similarity, and cross-modal dissimilarity for estimating robustness.
After placing \texttt{[MASK]} tokens in front of the identified objects mentions or similar nouns in $\mathcal{T}$, we consider the top-$k$ BERT predictions for each of the \texttt{[MASK]} words, denoted by $w_i$ $\forall$ $i \in \{1, \ldots, k\}$.   
We take the predicted probability scores of these $k$ words and normalize them to sum to one, denoting each by $p_i$.
The attribute similarity step computes  the similarities for $w_i$ with the corresponding attribute, which are then normalized to sum to one and denoted by $s_i$.  
Finally, we create $k$ augmentations of $\mathcal{T}$, each denoted by $\mathcal{T}'_i$, and compute the cosine distance of their CLIP embeddings with that of the corresponding image $\mathcal{I}$. The distances are also normalized to sum to one and denoted by $d_i$. Mathematically, the cumulative score for a predicted word $w_i$ is given as,
\begin{align}
\mathcal{S}_{w_i} = \lambda_1 \cdot p_i + \lambda_2 \cdot s_i + \lambda_3 \cdot d_i
\end{align}

where, $\lambda_1$, $\lambda_2$, and $\lambda_3$ are hyper-parameters that control the contribution of mask prediction using BERT, attribute similarity, and cross-modal dissimilarity, respectively. The word $w_i$ with the maximum score $\mathcal{S}$ is the word that is inserted in the place of the \texttt{[MASK]}. For text with multiple \texttt{[MASK]} tokens, we repeat this process iteratively in the order of their occurrence in $\mathcal{T}$.

By design, our cross-modal attribute insertion approach is modular as different components serve complementary functions toward the final objective of introducing semantically meaningful augmentations. It is also controllable using hyper-parameters $\lambda_1$, $\lambda_2$, $\lambda_3$, $k$, and $t$. Finally, our approach is training-free and, therefore, can be applied to investigate several tasks and models. 

\section{Experiments}
We study the effect of cross-modal attribute insertions on two multimodal tasks: text-to-image retrieval (i.e., retrieving images for a textual description) and cross-modal entailment (i.e., predicting the relationship between textual hypothesis and visual premise).

\subsection{Text $\rightarrow$ Image Retrieval}
\noindent\textbf{Task}: Given a set of text and image pairs as input, the goal is to retrieve the associated image for each text. The retrieval occurs for each text over a set of images, in our case we use a subset of 1000 text-image pairs, with the objective being to rank the original/ground-truth image the highest.\\
\textit{\underline{Axiom for Retrieval}}: Given  an image $\mathcal{I}$ in the search repository and two search queries $Q_1$ and $Q_2$, such that $Q_1$ contains more specific details of objects than $Q_2$, $\mathcal{I}$ should be retrieved at the same or higher rank for query $Q_1$ than for $Q_2$. 

\vspace{0.05in}
\noindent\textbf{Dataset}: For this task, we use the MSCOCO dataset~\cite{lin2014microsoft}, which contains images-caption pairs. Specifically, we use the image-caption pairs from 2017's validation split. This subset of the data contains $5,000$ unique images and $25,010$ captions, where each image can have multiple captions. To assess robustness, we perform augmentations on $25,010$ captions one by one while ranking all the images for each caption.

\vspace{0.05in}
\noindent\textbf{Model Under Investigation}: For this task we consider the CLIP model (\texttt{ViT-B/32}) \cite{radford2021learning}. CLIP is pretrained on 400 million image-caption pairs using contrastive learning, resulting in image and text encoders that produce unimodal embeddings that lie in a common latent space. 
The CLIP model has demonstrated great generalizability to various downstream tasks, including zero-shot and few-shot image classification and cross-modal retrieval. We obtain the CLIP embeddings for each image and caption in the MSCOCO dataset and rank all the images for a given caption based on their cosine similarities. We then contrast the ranking performance of the CLIP model using the original and augmented captions as textual queries. 

\subsection{Cross-Modal Entailment}
\noindent\textbf{Task}: Cross-modal entailment aims to determine whether the relationship between a visual premise and a textual hypothesis is `entailment,' `contradiction,' `neutral.' Specifically, `entailment' is observed when the textual hypothesis is logically implied (true) by the image while `contradiction' indicates that the textual hypothesis is not implied (false) by the visual premise. Finally, `neutral' represents an inconclusive or uncertain relationship between the hypothesis and the premise.\\
\textit{\underline{Axiom for Entailment}}: If the relationship between a visual premise and a textual hypothesis is `entailment,' it should not change to `contradictory' if the textual hypothesis is enriched with the information from the visual modality.

\vspace{0.05in}
\noindent\textbf{Dataset}: We perform this task on SNLI-VE~\cite{xie2019visual}, a visual entailment dataset. We use the test set of the dataset, comprising $17,859$ image (premise) \& text (hypothesis) pairs. For robustness assessment, we augment all text hypotheses while keeping the visual premise the same.

\vspace{0.05in}
\noindent\textbf{Model Under Investigation}: We investigate the pre-trained METER model \cite{dou2022empirical}, which consists of vision-and-language transformers that are trained end-to-end. The model's comprises CLIP's vision encoder and RoBERTa \cite{liu2019roberta} as the text encoder. The model pretraining objectives consist of masked language modeling and image-text-matching on four datasets: MSCOCO, Conceptual Captions~\cite{sharma-etal-2018-conceptual}, SBU Captions~\cite{ordonez2011im2text}, and Visual Genome~\cite{krishna2017visual}. In addition, METER is fine-tuned on SNLI-VE in order to achieve competitive performance. We contrast the performance of the model on the unmodified SNLI-VE dataset with the performance on the augmented version of SNLI-VE dataset.

\subsection{Baselines for Perturbations}
We compare our cross-modal attribute insertion approach (XMAI) with competitive baselines that are capable of introducing perturbations based on text-only information. We utilize the TextAttack~\cite{morris2020textattack}\footnote{\url{https://github.com/QData/TextAttack}} framework for implementing all the baseline perturbation strategies.

\noindent\textbf{Deletion}: A perturbation strategy that randomly removes words from the text.

\noindent\textbf{EDA} ~\cite{wei-zou-2019-eda}: This approach combines random deletion, random swapping, random insertion, and synonym replacement to modify each caption. We keep all parameters as default and set the percentage of words to swap to $20\%$. 

\noindent\textbf{CheckList} ~\cite{ribeiro-etal-2020-beyond}: Developed to generate a diverse set of evaluation examples, CheckList works by coalescing name replacement, location replacement, number alteration, and word contraction/extension.

\noindent\textbf{CLARE} ~\cite{li-etal-2021-contextualized}: This perturbation strategy uses language models to replace, insert, and merge tokens in captions. We use TextAttack's default fast implementation of CLARE.

\subsection{XMAI Implementation Details}
We choose $k=3$ for the number of top-k predicted BERT words for each \texttt{[MASK]} token and \texttt{flair/pos-english-fast} for PoS tagging of text. Next, to compare the nouns in the text with the objects identified in the image, we use word embeddings  produced by a Transformer-based model (\texttt{bert-base-nli-mean-tokens} on HuggingFace~\cite{wolf-etal-2020-transformers}). We set the threshold, $t$, for cosine similarity between nouns in $\mathcal{T}$ and objects in $\mathcal{I}$ to be $0.7$. For \texttt{[MASK]} filling, we use the \texttt{bert-base-cased} model on HuggingFace and the list of stopwords is adopted from NLTK.
\footnote{\url{https://www.nltk.org/}} To compute the similarity between attributes detected in $\mathcal{I}$ and BERT predictions, we employ SpaCy's pretrained tok2vec model (\texttt{en\_core\_web\_md}), which contains 300-dimensional embeddings for $\sim500 k$ words~\cite{honnibal2020spacy}.  Lastly, the pre-trained CLIP model (\texttt{ViT-B/32}) is used to compute image and text embeddings in a common latent space. For our main experiments, we set the values of $\lambda_i$ as $\lambda_1 = 1$, $\lambda_2 = 5$, and $\lambda_3 = 5$.

\begin{table*}[!t]
\centering
\resizebox{0.7\textwidth}{!}{%
\begin{tabular}{lccccc}
\hline
\textbf{Approach} & \textbf{MRR} $\downarrow$ & \textbf{$\boldsymbol{\mathcal{S}im_{\mathcal{T}-\mathcal{T}'}}$} $\uparrow$ & \textbf{$\boldsymbol{\mathcal{S}im_{\mathcal{I}-\mathcal{T}'}}$} $\uparrow$ & \textbf{BLEU} $\uparrow$ & \textbf{METEOR} $\uparrow$\\
\hline
Original & 0.632 & 1.000 & 0.294 & 1.000 & 1.000\\\hline
Deletion & 0.581 & \underline{0.948} & \underline{0.288} & \underline{0.758} & 0.910\\
EDA & 0.570 & 0.914 & 0.287 & 0.647 & 0.931\\
Checklist & 0.630 & \textbf{0.996} & \textbf{0.294} & \textbf{0.982} & \textbf{0.993}\\
CLARE & \underline{0.567} & 0.899 & 0.285 & 0.749 & 0.947\\
\hline
\textbf{XMAI (Ours)} & \textbf{0.536} & 0.924 & 0.283 & 0.623 & \underline{0.969}\\
\hline
\end{tabular}
}
\caption{{
\textbf{Results on the text-to-image retrieval task.} The effectiveness of augmentation in highlighting the model's vulnerability is noted by the drop in MRR with respect to the original MRR score. $\mathcal{S}im_{\mathcal{T}-\mathcal{T}'}$, BLEU, and METEOR capture the relevance of augmented text with the original text and $\mathcal{S}im_{\mathcal{I}-\mathcal{T}'}$ captures the relevance of the augmented text with the original text. Best results are in \textbf{bold}; second best are \underline{underlined}.}
}

\label{tab:t2i_retrieval}
\end{table*}

\begin{table*}[!t]
\centering
\resizebox{2.0\columnwidth}{!}{%
\begin{tabular}{lcccccccc}
\hline
\textbf{Approach} & \textbf{Acc.} $\downarrow$ & \textbf{Precision} $\downarrow$ & \textbf{Recall} $\downarrow$ & \textbf{$\mathbf{F_1}$} $\downarrow$ & \textbf{$\boldsymbol{\mathcal{S}im_{\mathcal{T}-\mathcal{T}'}}$} $\uparrow$ & \textbf{$\boldsymbol{\mathcal{S}im_{\mathcal{I}-\mathcal{T}'}}$} $\uparrow$ & \textbf{BLEU} $\uparrow$ & \textbf{METEOR} $\uparrow$\\
\hline
Original & 0.792 & 0.790 & 0.792 & 0.791 & 1.000 & 0.246 & 1.000 & 0.998\\
\hline
Deletion & 0.742 & 0.742 & 0.742 & 0.740 & \underline{0.892} & 0.240 & \underline{0.632} & 0.853\\
EDA & 0.719 & 0.719 & 0.719 & 0.718 & 0.878 & \underline{0.241} & 0.541 & 0.900\\
Checklist & 0.791 & 0.790 & 0.791 & 0.790 & \textbf{0.973} & \textbf{0.246} & \textbf{0.993} & \textbf{0.986}\\
CLARE & \textbf{0.632} & \textbf{0.652} & \textbf{0.631} & \textbf{0.618} & 0.804 & 0.238 & 0.592 & 0.911\\
\hline
\textbf{XMAI (Ours)} & \underline{0.643} & \underline{0.682} & \underline{0.643} & \underline{0.625} & 0.873 & 0.235 & 0.621 & \underline{0.963}\\
\hline
\end{tabular}
}
\caption{{
\textbf{Results on the cross-modal entailment task.} Augmentations that cause a greater drop in classification metrics are more effective at highlighting the lack of multimodal robustness, while the similarity metrics capture their relevance with the original example. The best results are in \textbf{bold} and the second best results are \underline{underlined}.}
}
\label{tab:visual_ent}
\end{table*}

\subsection{Evaluation Metrics}
We measure the impact of perturbations in the text on the capabilities of multimodal models using task-specific metrics. We quantify text-to-image retrieval performance using mean reciprocal rank (MRR). For cross-modal entailment, we report standard classification metrics (accuracy, precision, recall, and $F_1$ score). 

While the effectiveness of perturbations is important for highlighting model vulnerabilities, it is also imperative to measure the relevance of augmented text $\mathcal{T}'$ with original text $\mathcal{T}$ and image $\mathcal{I}$. To this end, we compute mean cosine similarity $\mathcal{S}im_{\mathcal{T}-\mathcal{T}'}$ between original and modified texts (i.e., $\mathcal{T}$ \& $\mathcal{T}'$, respectively) using a sentence Transformer model (\texttt{all-mpnet-base-v2}) ~\cite{reimers2019sentence}. Similarly, we report BLEU~\cite{papineni-etal-2002-bleu} and METEOR~\cite{banerjee2005meteor}, using NLTK to further compare the texts (considering n-grams of size upto $4$ in the case of BLEU).
    Additionally, we compute the mean cosine similarity $\mathcal{S}im_{\mathcal{I}-\mathcal{T}'}$ using CLIP (\texttt{ViT-B/32}) embeddings.

\section{Results and Analysis}
Recall that our primary goal is to use XMAI to obtain a complementary and novel set of text augmentations that can highlight the vulnerabilities of multimodal models. To this end, we contrast the performance of the models under investigation on the original and the modified examples, and quantify the relevance of the modified text with respect to the original text and image. We recruit human annotators to compare the quality of the augmentations generated using our approach with (i) the ones generated using the most competitive baseline, and (ii) the original text. 

\begin{figure*}[!h]
    \centering
    \includegraphics[width=1.0\linewidth]{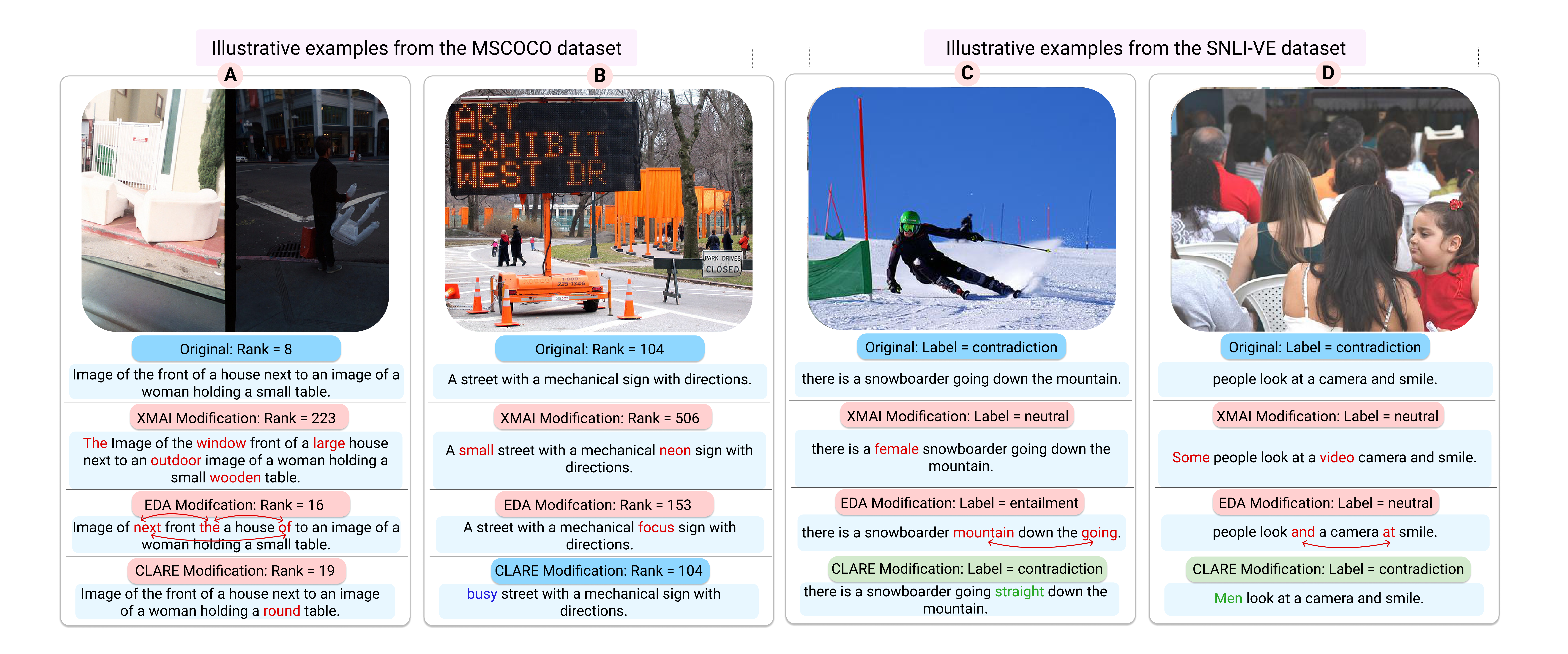}\vspace{-3mm}
    \caption{{Qualitative examples comparing the augmentations produced by our XMAI method to both EDA and CLARE on both MSCOCO and SNLI-VE tasks. Red text represents a drop in rank or misclassification, green text indicates improvement in rank or correct classification, and blue marks when a change has no impact on rank. Lastly, arrows and the words at either end of each arrow indicate swapping by EDA.}}
    \label{fig:qual-examples}
\end{figure*}

\vspace{0.05in}
\noindent\textbf{Robustness of Multimodal Models}: Table \ref{tab:t2i_retrieval} shows that for the text $\rightarrow$ image retrieval task, our cross-modal attribute insertions cause the greatest drop in observed MRR; the MRR drops from an original value of $0.632$ to $0.536$. Similarly, Table \ref{tab:visual_ent} shows that for the cross-modal entailment task our approach performs second only to CLARE --- an observation that is consistent across all the metrics, accuracy, precision, recall, and $F_1$. It is worth noting that our approach is the only one that uses the information from the image modality to introduce textual perturbations and hence the resultant perturbed examples are characteristically different than the ones that are produced using baseline methods like CLARE. We will revisit this using qualitative examples. Overall, results demonstrate that  state-of-the-art vision-and-language learning approaches for text-to-image retrieval and cross-modal entailment tasks are not robust to our proposed augmentations.

\vspace{0.05in}
\noindent\textbf{Relevance of Augmentations}: Tables \ref{tab:t2i_retrieval} and \ref{tab:visual_ent} show that XMAI produces augmentations $\mathcal{T}'$ that maintain high-relevance with the original text $\mathcal{T}$ and the image $\mathcal{I}$, in terms of $\mathcal{S}im_{\mathcal{T}-\mathcal{T}'}$ and $\mathcal{S}im_{\mathcal{I}-\mathcal{T}'}$.
It is interesting to note that the BLEU scores for augmentations generated by XMAI are notably lower than that for the baseline augmentations. On the contrary, METEOR scores show that XMAI's augmentations are "closer" to the original texts compared to most baselines. XMAI's poor BLEU scores can be largely attributed to BLEU's tendency to penalize novel insertions severely compared to removals or replacements, as it is a precision-based metric~\cite{banerjee2005meteor}.\footnote{Novel insertions in $\mathcal{T}'$ mean more `false positives' with respect to $\mathcal{T}$, indicating lower precision and BLEU scores.} In Appendix \ref{sec:app_num_insertions} (Table \ref{tab:num-perturbs}), we further note that, on average, XMAI inserts $1.660$ $(\pm 0.947)$ new words in MSCOCO captions and $1.269$ $(\pm 0.768)$ new words in SNLI-VE hypotheses. This is considerably higher than the rate of insertions made by other methods, especially Checklist, where an obvious consequence of making a fewer number of augmentations is better text similarity across a corpus. We thus attribute the poor performance of XMAI in terms of BLEU scores to BLEU's inability to handle insertions appropriately. This is further substantiated by the human assessments.

\subsection{Human Assessment of Augmentations}
\label{sec:human_assessment}
We recruit annotators using Amazon Mechanical Turk (AMT) to answer the following two questions: \textit{(i)} do cross-modal attribute insertions lead to better text augmentations than the most competitive baseline (i.e., CLARE), and \textit{(ii)} are cross-modal attribute insertions as good as the original text accompanying the image. Please see Appendix \ref{sec:human_eval} for details on the recruitment filters and compensation on AMT. 

\noindent\textbf{XMAI versus CLARE}: 
We randomly sampled 100 examples from the validation set of the MSCOCO dataset and showed the modified captions using CLARE and XMAI. $5$ annotators annotated each example. We asked annotators to indicate their agreement to the following question after seeing two captions for a given image using a 5-point Likert scale (1: Strongly disagree, ..., 5: Strongly agree): \textit{\texttt{Caption 2} is a better description of the shown image than \texttt{Caption 1} in terms of its quality and accuracy}. \texttt{Caption 1} and \texttt{Caption 2} were randomly flipped between CLARE and XMAI to avoid any position bias. Furthermore, to ensure quality annotations, we randomly inserted some ``attention check'' examples that instructed annotators to ignore all previous instructions and mark specified responses on the Likert scale. We discarded responses from annotators who marked the attention-check examples incorrectly and re-collected the annotations.

For $63\%$ of the examples, a majority of annotators (i.e., at least $3$ out of $5$) preferred the captions modified using XMAI over CLARE. The captions modified using CLARE were preferred for $26\%$ examples. The rest were either marked as `equivalent' (i.e., 3: Neither disagree nor agree) or had ambiguous majority votes. Overall, the results demonstrate that the annotators preferred the captions modified using XMAI over the ones modified using CLARE, in terms of their accuracy in describing the image and their quality. We next assess how XMAI modified captions compare against the original captions.

\begin{figure*}
    \centering
    \includegraphics[width=0.9\linewidth]{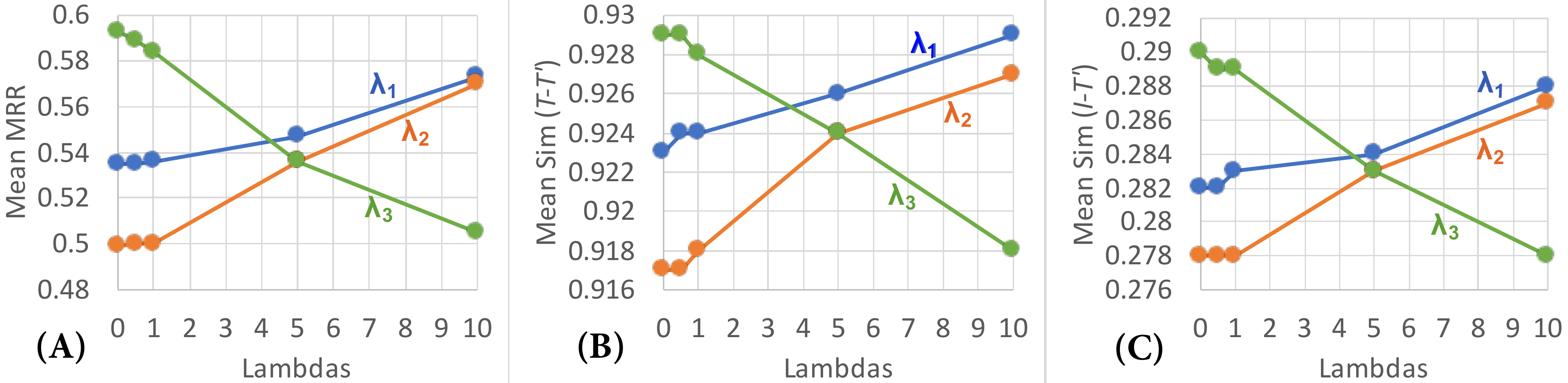} 
    \caption{{\textbf{Varying $\lambda_i$ to isolate their effect on the text-to-image retrieval task.} Ablations on independent effects of lambda values, where the default lambdas are: $\lambda_1 = 1$, $\lambda_2 = 5$, and $\lambda_3 = 5$. Each line plot represents changing the specified $\lambda$ while keeping the others as default. We observe the variation in task-specific performance as well as the similarity metrics.}}
    \label{fig:lambda_ablations_mscoco}
\end{figure*}

\noindent\textbf{XMAI versus Original}: We randomly sampled 100 examples from the validation set of the MSCOCO dataset and randomly chose $50$ of them to be modified using XMAI while leaving the other $50$ unmodified. We first primed the annotators to view $5$ original image caption pairs, noting them as reference examples.\footnote{These $5$ reference examples were not included in the subset of 100 examples selected for annotations.} We then asked the annotators to view a \textit{list} of image-caption pairs and evaluate the caption quality using the following prompt: \textit{Rate the caption quality for the given image based on the reference examples shown earlier}. A response of $1$ on the 5-point Likert scale indicated `extremely poor quality' whereas that of $5$ indicated `extremely good quality.' The shown list comprised randomly shuffled original image-caption pairs and modified image-caption pairs using XMAI, and a few attention-check examples. Each example received annotations from $5$ different annotators.

The unmodified captions received an average score of $4.12$ $(\pm 0.37)$ whereas that for the modified caption using XMAI was $4.07$ $(\pm 0.33)$. The observed inter-rater agreement was strong, with a \ Krippendorf's $\alpha$ of $0.78$. Additionally, a two-sided t-test with unequal variances assumption failed to reject the null hypothesis ($ p > 0.05$) that the average Likert scores for the original and modified captions are from different distributions. In sum, the perceived quality of the modified captions using XMAI is not statistically different from that of the original captions.

\noindent\textbf{Computational Efficiency}: In Appendix Fig. \ref{fig:perturbation_times} we demonstrate that our approach for inserting cross-modal attributes is $14.8\times$ and $9.4\times$ faster than the most competitive baseline approach (i.e., CLARE) on MSCOCO and SNLI-VE, respectively. Combined with the fact that XMAI augmentations are perceived to be of better quality than CLARE augmentations and are effective at highlighting model vulnerabilities, the increased computational efficiency allows for more rapid and realistic model validation. In the following section, we demonstrate via qualitative examples that, being the only approach that leverages cross-modal interactions in multimodal data, the augmentations produced by XMAI are novel to those produced by the text-only baselines. 

\subsection{Qualitative Analysis}

In Figure \ref{fig:qual-examples} we show illustrative examples of the insertions introduced using our approach and contrast them with existing text-only perturbations. We observe that our cross-modal insertions lead to a complementary set of augmentations that are not covered by text-only approaches.

We note that our method does not remove any information present in the original caption/hypothesis. This prevents our method from drastically changing the original semantics, which has been a known shortcoming of text-only perturbations~\cite{wang2021adversarial}. In Figure \ref{fig:qual-examples}(A), we note that EDA produces a grammatically incoherent augmentation (``\textit{Image of next front the a house of to...}") and CLARE inserts an inaccurate attribute (``\textit{round table}"). Whereas, our approach only inserts relevant attributes to the original text (``\textit{The Image of the window front of a large house next to an outdoor image of a woman holding a small wooden table.}"). In Figure \ref{fig:qual-examples}(B\&D) we see that XMAI modifies the text using the information in the corresponding images --- for instance, our approach identifies the neon LEDs and inserts `neon' in front of `sign.' However, EDA and CLARE introduce inaccurate details. XMAI is also capable of multiple meaningful insertions. Our work is the first to enable cross-modal insertion capabilities to obtain meaningful augmentations of multimodal (image + text) data.

\subsection{Ablations for $\lambda_i$ Sensitivity}
 In Figure \ref{fig:lambda_ablations_mscoco}, we visualize the change in retrieval performance with respect to independent changes in $\lambda_1$, $\lambda_2$, and $\lambda_3$. In other words, we vary a given $\lambda_i$ while keeping other hyper-parameters at their aforementioned values. 
We find that increasing $\lambda_1$ and $\lambda_2$ improves the relevance of augmentations but reduces their effectiveness in highlighting vulnerabilities. Intuitively, these components increase the likelihood that our approach picks insertions with high BERT prediction scores (controlled by $\lambda_1$) and similarities with the identified image attribute (controlled by $\lambda_2$). On the other hand, increasing $\lambda_3$, which controls the contributions of the robustness assessment component, generates less relevant augmentations that are highly effective. This observation also aligns with our goal of exploiting the lack of robust encoding mechanisms to highlight model vulnerabilities. 

Overall, these results demonstrate that the individual components of our approach play significant roles, and can be controlled using the $\lambda_i$ hyper-parameters. Similar trends are observed for the cross-modal entailment task; see Appendix Fig. \ref{fig:lambda_ablations_snli}. We discuss the ablations pertaining to the similarity threshold for matching image objects and text nouns in Appendix \ref{sec:t_ablations}.

\section{Conclusion}
A \textit{robust} understanding of vision-and-language data is crucial for powering several applications. We propose cross-modal attribute insertions – i.e., adding the visual attributes to text, as a new variation that is likely in multimodal data and to which multimodal models should be robust. Our approach produces novel augmentations that are complementary to existing methods that model text-only data, and are preferred over them by human annotators. Using our augmentations we effectively highlight the vulnerabilities of state-of-the-art multimodal models for text-to-image retrieval and cross-modal entailment. In the future, we aim to empirically study the effect of including XMAI augmented data in task-specific training sets and expand to a broader set of multimodal tasks and metrics.

\section{Limitations and Broader Perspective}
\noindent\textit{Limitations and bias of pre-trained models}: Our work uses detected objects and their attributes in the images to introduce novel insertions in the corresponding text. To this end, it is important to address the limitations of the state-of-the-art object and attribute detection methods. The undesired artifacts of these methods could be categorized as inaccurate or biased. The detected objects could be incorrect, but since we only consider objects that are  also mentioned in the text, the effect of incorrect object detections is non-existent in our augmentations. However, we notice that some of the detected attributes in images and BERT predictions reflect stereotypical associations and have been documented in prior works~\cite{li2021discover, kaneko2022unmasking}. We acknowledge that the current state of deep learning research is limited, and the consequential shortcomings are reflected in our augmentations to some extent.

\vspace{0.01in}
\noindent\textit{Broader social impact}: The authors do not foresee any negative social impacts of this work. We believe our cross-modal augmentations will enable an exhaustive evaluation of the robustness of vision-and-language models, leading to more reliable multimodal systems. We release the code for our experiments to aid reproducibility and enable future research on this topic.

\vspace{0.01in}\noindent\textit{Annotations, IRB approval, and datasets}: The annotators for evaluations done in this study were recruited via Amazon Mechanical Turk. We specifically recruited `Master' annotators located in the United States; and paid them at an hourly rate of $12$ USD for their annotations. The human evaluation experiments were approved by the Institutional Review Board (IRB) at the authors' institution. The datasets used in this study are publicly available and were curated by previous research. We abide by their terms of use.

\section{Acknowledgements}
This research/material is based upon work supported in part by NSF grants 
CNS-2154118, IIS-2027689, ITE-2137724, ITE-2230692, CNS-2239879, Defense Advanced Research
Projects Agency (DARPA) under Agreement No.
HR00112290102 (subcontract No. PO70745), and funding from Microsoft, Google, and Adobe Inc. GV is partly supported by the Snap Research Fellowship. Any opinions, findings, and conclusions or recommendations expressed in this material are those of the author(s) and do not necessarily reflect the position or policy
of DARPA, DoD, SRI International, NSF and no official endorsement should be inferred. We thank the anonymous reviewers for their constructive comments and the CLAWS research group members for their help with the project.

\bibliography{anthology,custom}
\bibliographystyle{acl_natbib}

\appendix

\section{Appendix}
\label{sec:appendix}

\begin{figure*}
\centering
\begin{subfigure}{0.45\linewidth}
  \centering
  \includegraphics[width=1.0\linewidth]{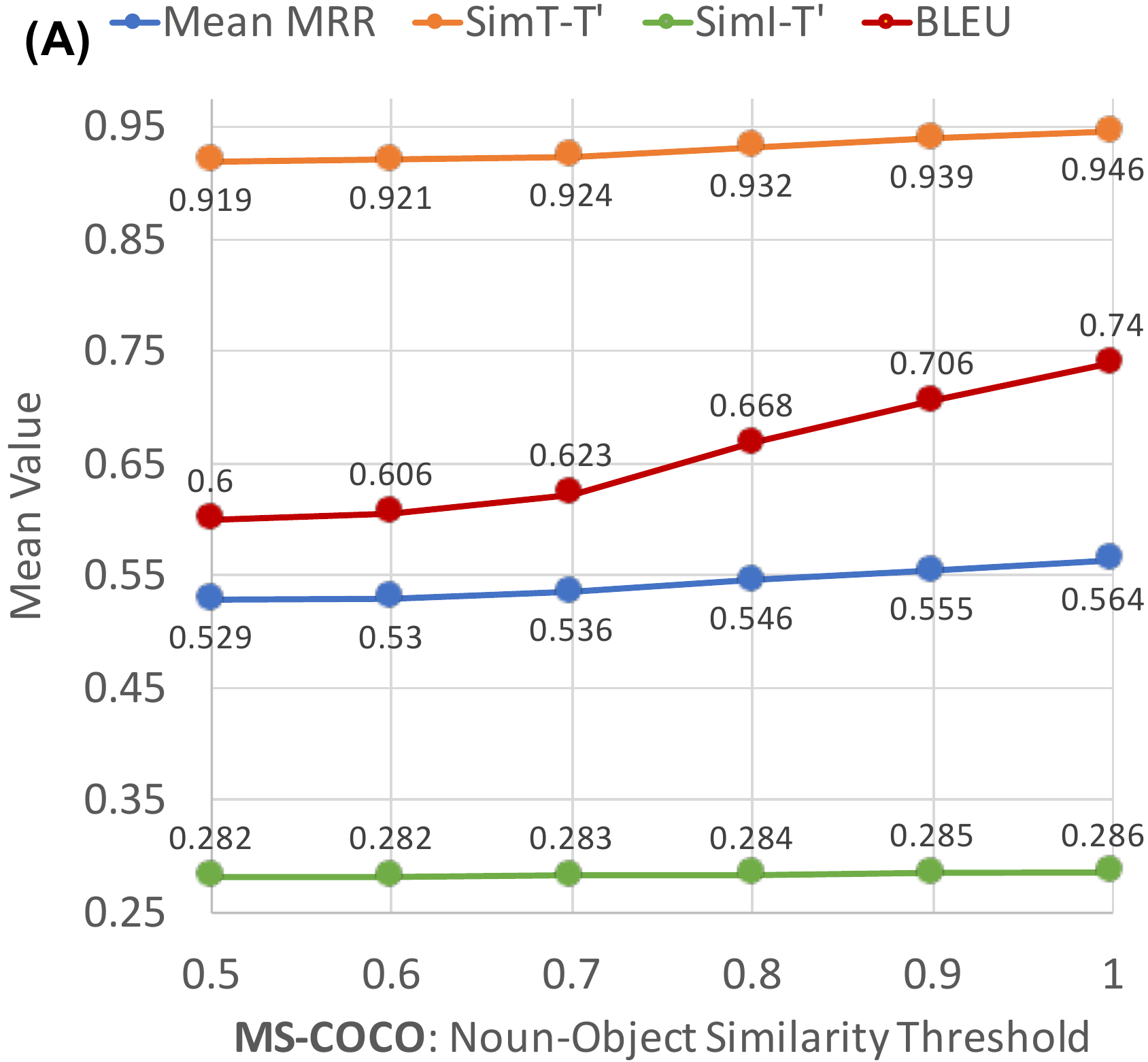}
  \caption{Ablation on the noun-object similarity threshold $t$ for the text-to-image retrieval task (MSCOCO).}
  \label{fig:t_coco_ablations}
\end{subfigure}\hspace{12pt}
\begin{subfigure}{0.43\linewidth}
  \centering
  \includegraphics[width=1.0\linewidth]{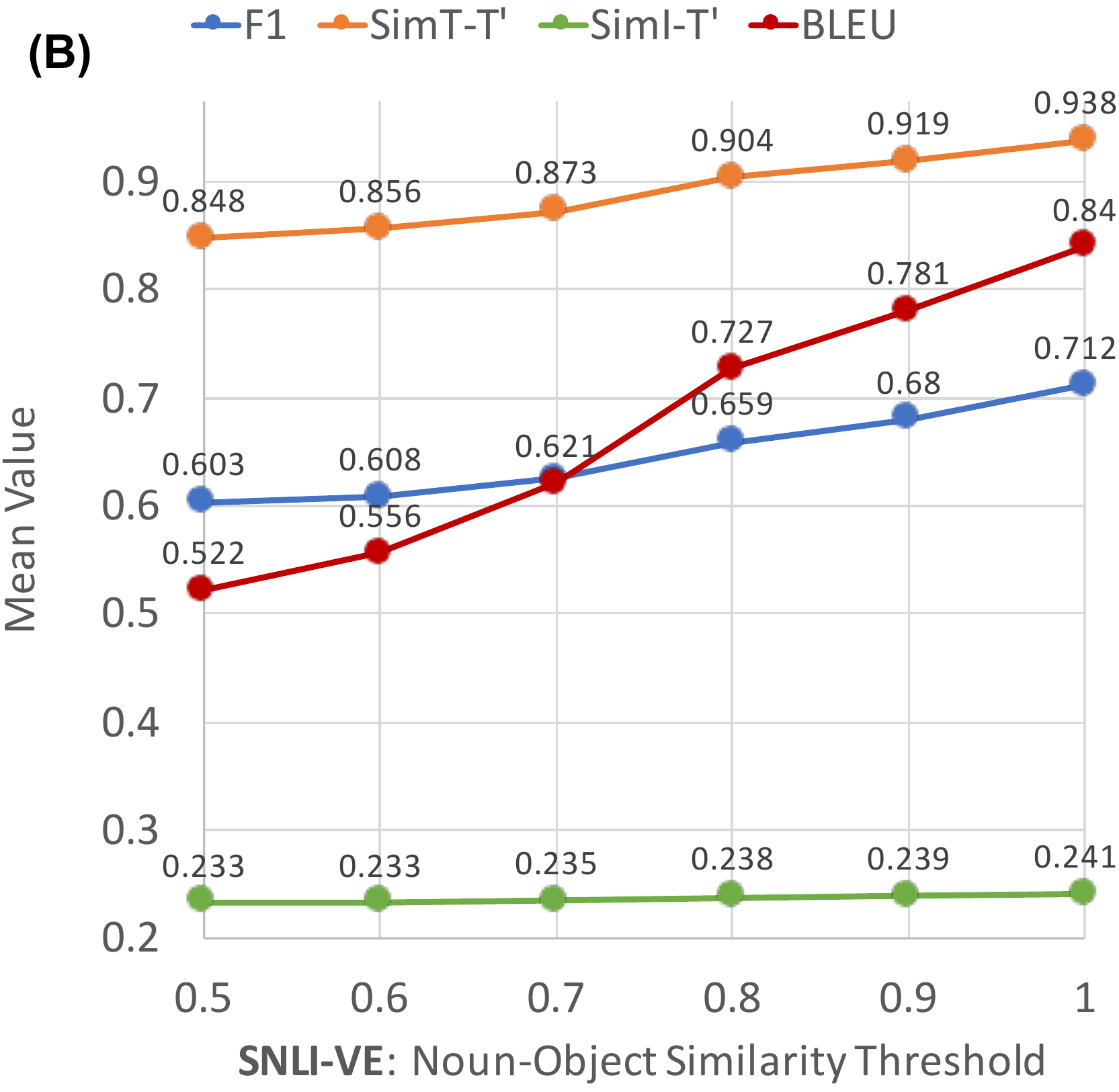}
  \caption{Ablation on the noun-object similarity threshold $t$ for the cross-modal entailment task (SNLI-VE).}
  \label{fig:t_snli_ablations}
\end{subfigure}
\caption{Ablations on threshold, $t$, across task-specific metrics for both tasks.}
\label{fig:t_ablations}
\end{figure*}

\subsection{Human Evaluation Details}
\label{sec:human_eval}
For our annotation tasks, we recruited annotators using Amazon Mechanical Turk. We set the criteria to `Master' annotators who had at least $99\%$ approval rate and were located in the United States. The rewards were set by assuming an hourly rate of 12 USD for all the annotators. In addition, the annotators were informed that the aggregate statistics of their annotations would be used and shared as part of academic research. 

Previous research has demonstrated the role of providing priming examples in obtaining high-quality annotations~\cite{DBLP:journals/corr/abs-2101-06561}. Therefore, we showed unmodified examples from the MSCOCO corpus to help annotators establish a reference for quality and accuracy. 
For both the crowd-sourced evaluations, we inserted some ``attention-check'' examples to ensure the annotators read the text carefully before responding. This was done by explicitly asking the annotators to mark a randomly-chosen score on the Likert scale regardless of the actual content. We discard the annotations from annotators who did not correctly respond to all the attention-check examples.

\subsection{Further Ablations}
\label{sec:t_ablations}

\noindent\textbf{Extended Variations in $t$}: To illustrate the effect of changing the threshold, $t$, we plot our described MSCOCO metrics with respect to variations in $t$ in Figure \ref{fig:t_ablations}. We see that as the criterion for matching nouns in the text and objects in the image is made more stringent, the quality and relevance of the augmentations improve further. However, the effectiveness of the resulting augmentations in highlighting model vulnerabilities decreases.  It is worth noting as the model becomes more selective in inserting attributes (due to fewer matched nouns and objects), we witness a stark increase in BLUE scores. Variations in $t$ effectively capture the trade-off between maintaining the relevance of augmentations and effectively highlighting model vulnerabilities. Even though it is possible to construct augmentations that will be more effective in making the multimodal models perform poorly,  it would sacrifice the relevance and quality of resulting augmentations. Our main results demonstrate that with $t=0.7$, we obtain high-quality and human-preferred augmentations that are also effective in highlighting vulnerabilities. 

\begin{table*}[!t]
\centering
\resizebox{0.7\linewidth}{!}{%
\begin{tabular}{ccccccc}
% \toprule
\textbf{$\boldsymbol{k}$} & \textbf{MRR} $\downarrow$ & \textbf{$\boldsymbol{\mathcal{S}im_{\mathcal{T}-\mathcal{T}'}}$} $\uparrow$ & \textbf{$\boldsymbol{\mathcal{S}im_{\mathcal{I}-\mathcal{T}'}}$} $\uparrow$ & \textbf{BLEU} $\uparrow$ & \textbf{METEOR} $\uparrow$ & \textbf{Time (s)} $\downarrow$ \\
\hline
 3 & 0.536 & 0.924 & 0.283 & 0.623 & 0.969 & 0.196\\

 5 & 0.506 & 0.919 & 0.278 & 0.623 & 0.970 & 0.311 \\

 10 & 0.476 & 0.914 & 0.274 & 0.623 & 0.970 & 0.474 \\
\hline
\end{tabular}
}
\caption{\label{tab:mixed_ablation_retrieval}
Ablations by varying the number of BERT predictions considered (i.e., $k$) for the text-to-image retrieval task on MSCOCO. The reported time is in seconds (per caption). 
}
\end{table*}

\begin{table*}[!t]
\centering
\resizebox{0.7\linewidth}{!}{%
\begin{tabular}[width = \linewidth]{c c  | c  }
\textbf{Method} & \textbf{MSCOCO Captions} & \textbf{SNLI-VE Hypotheses}\\
\hline
 & $\mu$ ($\pm \sigma$) & $\mu$ ($\pm \sigma$) \\
\hline
Deletion & 0.000 ($\pm$ 0.022) & 0.000 ($\pm$ 0.015)\\
EDA & 0.679	($\pm$ 0.784) & 0.562 ($\pm$ 0.633)\\
Checklist & 0.077 ($\pm$ 0.284) & 0.087 ($\pm$ 0.300) \\
CLARE & 0.982 ($\pm$ 0.161) &  0.990 ($\pm$ 0.132) \\
\hline
\textbf{XMAI (Ours)} & 1.660 ($\pm$ 0.947) & 1.269 ($\pm$ 0.768) \\
\hline
\end{tabular}
}
\caption{\label{tab:num-perturbs}
Mean (and standard deviation) of the number of novel insertions/replacements in modified MSCOCO captions and SNLI-VE hypotheses with respect to their original counterparts.
}
\end{table*}

\begin{table*}[!t]
\centering
\resizebox{0.7\linewidth}{!}{%
\begin{tabular}[width = \linewidth]{c r|r|r|r  }
\textbf{Method} & \multicolumn{2}{c}{\textbf{MSCOCO Captions}} & \multicolumn{2}{c}{\textbf{SNLI-VE Hypotheses}}\\
\hline
 & \textit{r/i} & \textit{any} & \textit{r/i} & \textit{any}\\
\hline
Deletion & $12$ & $25,010$ & $4$ & $17,857$\\
EDA & $12,492$ & $24,980$ & $8,876$ & $17,786$\\
Checklist & $1,817$ & $1,879$ & $1,464$ & $1,492$\\
CLARE & $24,456$ & $24,972$ & $17,619$ & $17,817$\\
\hline
\textbf{XMAI (Ours)} & $24,244$ & $24,970$ & $16,275$ & $16,512$\\
\hline
\end{tabular}
}
\caption{\label{tab:num-captions}
Number of examples that are modified in MSCOCO and SNLI-VE. As mentioned in the Experiments section, we consider 25,010 captions for MSCOCO and 17,859 hypotheses for SNLI-VE. We further split this by considering modifications as only novel replacements/insertions (\textit{r/i}) or any difference between original and modified (\textit{any}).
}
\end{table*}

\begin{table*}[!h]
\centering
\resizebox{1.0\linewidth}{!}{%
\begin{tabular}{ccccccccccc}
\hline
\textbf{Ablation} & \textbf{$\boldsymbol{\lambda_1}$} & \textbf{$\boldsymbol{\lambda_2}$} & \textbf{$\boldsymbol{\lambda_3}$} & \textbf{$\boldsymbol{k}$} & \textbf{$\boldsymbol{t}$} & \textbf{MRR} $\downarrow$ & \textbf{$\boldsymbol{\mathcal{S}im_{\mathcal{T}-\mathcal{T}'}}$} $\uparrow$ & \textbf{$\boldsymbol{\mathcal{S}im_{\mathcal{I}-\mathcal{T}'}}$} $\uparrow$ & \textbf{BLEU} $\uparrow$ & \textbf{METEOR} $\uparrow$\\
\hline
Original
& 1 & 5 & 5 & 3 & 0.7 & 0.536 & 0.924 & 0.283 & 0.623 & 0.969\\
\hline
$\lambda_2$
& 0 & 1 & 0 & 3 & 0.7 & 0.593 & 0.929 & 0.290 & 0.623 & 0.970\\
& 0 & -1 & 0 & 3 & 0.7 & 0.588 & 0.922 & 0.290 & 0.623 & 0.968 \\
\hline
$\lambda_3$
& 0 & 0 & 1 & 3 & 0.7 & 0.499 & 0.917 & 0.277 & 0.623 & 0.969\\
& 0 & 0 & -1 & 3 & 0.7 & 0.672 & 0.932 & 0.301 & 0.623 & 0.969\\
\hline
\end{tabular}%
}
\caption{\label{tab:bc_ablation_retrieval}
Effect of variations in $\lambda_2$, $\lambda_3$, and $k$ for the text-to-image retrieval on MSCOCO.
}
\end{table*}

\begin{figure*}[!h]
    \centering
    \includegraphics[width=1.0\linewidth]{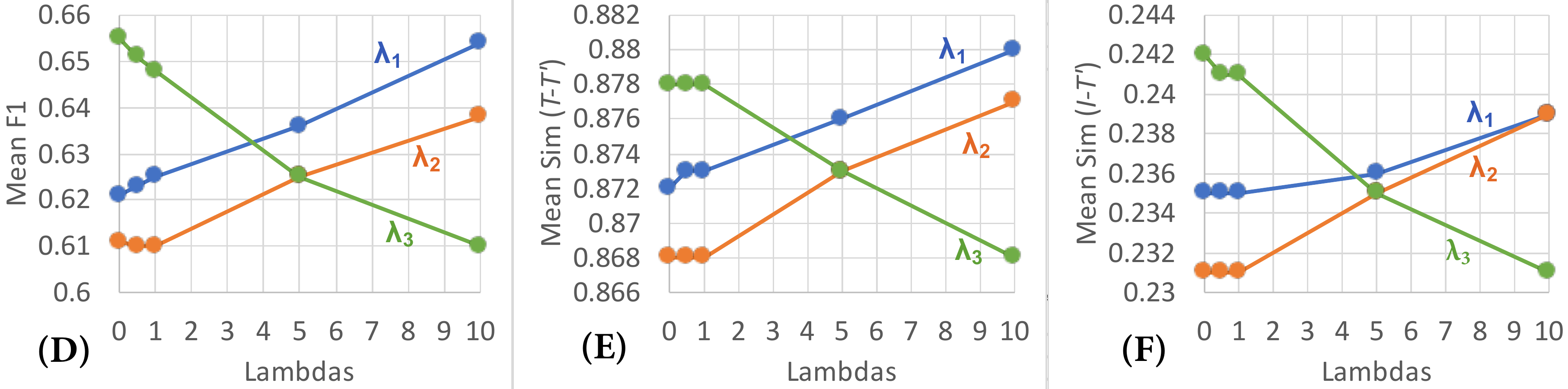}
\caption{{\textbf{Varying $\lambda_i$ to isolate their effect on the cross-mdodal entailment task.} Ablations on independent effects of lambda values, where the default lambdas are: $\lambda_1 = 1$, $\lambda_2 = 5$, and $\lambda_3 = 5$. Each line plot represents changing the specified $\lambda$ while keeping the others as default. We observe the variation in task-specific performance as well as the similarity metrics.}}
\label{fig:lambda_ablations_snli}
\end{figure*}

\vspace{0.05in}
\noindent\textbf{Variations in $k$}: We perform another ablation by increasing the value of $k$ for the top-k predictions made by pre-trained BERT model. Table \ref{tab:mixed_ablation_retrieval} shows that increasing the search space for possible insertion tokens leads to a notable drop in the retrieval performance over resulting augmentations. However, the relevance values with the original image and text drop too. Increasing the search space allows the model to explore potential insertions that could produce highly dissimilar cross-modal representations, thereby helping the adversarial component of our framework, but with the compromise being the relevancy of the augmentations. We also note that exploring more insertion possibilities increases the per-caption augmentation time taken by XMAI. 

\vspace{0.05in}
\noindent\textbf{Varying $\lambda_2$ and $\lambda_3$}: To comprehensively understand the effect of variations in $\lambda_i$, specifically $\lambda_2$ and $\lambda_3$, we set each to $1$ or $-1$ (one at a time) while setting the other two lambdas to $0$. Results in Table \ref{tab:mixed_ablation_retrieval} empirically show that both the attribute similarity and robustness assessment components are essential and can serve dual purposes; i.e., negative values of $\lambda_2$ and $\lambda_3$ allow the associated components to serve the opposite purpose. In contrast to their original objective, when the associated $\lambda$ values are set to negative values, attribute similarity decreases performance and robustness assessment increases it.

\subsection{Number of Insertions}
\label{sec:app_num_insertions}
We report the number of insertions or replacements each augmentation method makes to the original text as well as the number of texts modified for each dataset. 
The results are reported in Table \ref{tab:num-perturbs} and \ref{tab:num-captions}.
We find that our XMAI approach introduces more novel words in the augmentation than any other approach, while also augmenting nearly the same amount of captions as most competitive baseline approaches. This observation, combined with the fact that human annotators prefer XMAI augmentations over baseline augmentations, shows that cross-modal insertions can be used to introduce new information without causing semantic deterioration of the text. Additionally, these results allow us to attribute the low BLEU scores observed in Table \ref{tab:t2i_retrieval} to the higher number of insertions that XMAI makes. Note that BLEU computation is precision-based and hence penalizes novel insertions more severely.
It is interesting to note that even though `Deletion' is expected to have no insertions or replacements, we found that in very few cases, due to the adopted implementation as well as the noise in the text could result in fragmented or fused words that were being considered novel compared to the original text and therefore counted as an insertion/replacement.

\begin{figure}[!t]
    \centering
    \includegraphics[scale=0.4]{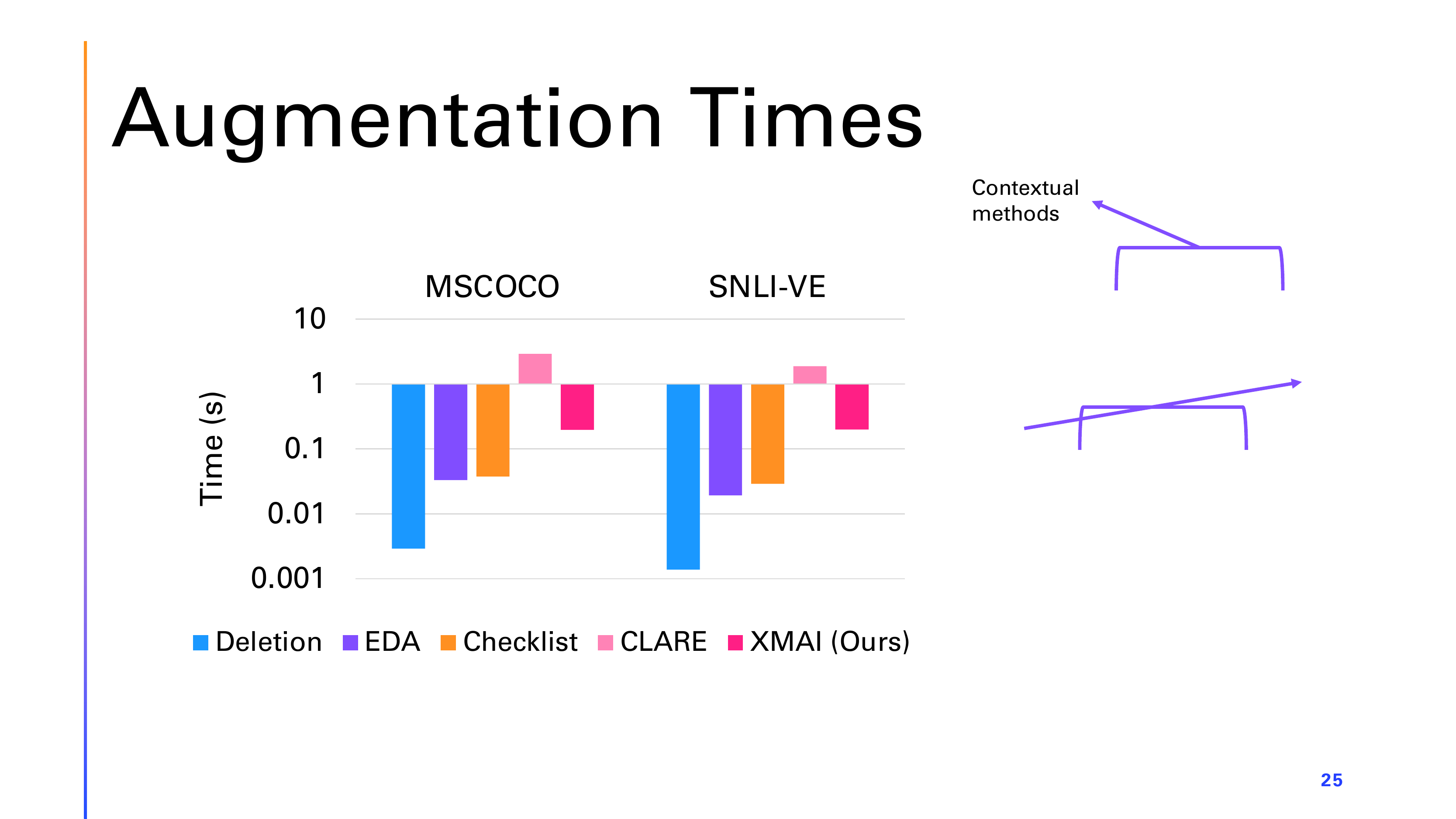}
    \caption{Comparing the per-caption augmentation time of various methods across the two tasks. Results are shown in logarithmic scale due to the disparity between computational times across different methods.}
    \label{fig:perturbation_times}
\end{figure}

\subsection{Augmentation Time}
In Figure \ref{fig:perturbation_times} we show the average time to augment the input text for each of the methods. The results are plotted using a logarithmic scale to ensure a clearer depiction across methods that vary considerably in terms of computational time.

Simpler approaches such as Deletion, EDA, and CheckList can modify tens or hundreds of samples each second. Intuitively, the reliance on simple rules makes these approaches very efficient. On the other hand, context-aware methods such as CLARE and XMAI are slower due to their reliance on large language models and involved selection processes. However, XMAI can augment the text a magnitude faster than CLARE, even after using the fast implementation of CLARE from TextAttack.

We don't consider the time to compute objects and attributes for XMAI for two reasons. First, the cost to perform this step was $< 1$ hour for our datasets and the relationship between methods remains the same. Secondly, objects and attributes only need to be computed once, so there is no additive cost for augmentation unless changes are made to the detection component.

\subsection{Compute Resources}
Our experiments were split between a single Tesla V100 for object detection ($\sim 1$ hour) and NVIDIA Tesla T4 GPUs for our augmentation ($\sim 3$ hours).

\begin{algorithm}
\caption{Algorithmic block describing the text augmentation method for XMAI. For details reference back to and follow along with Section \ref{sec:XMAI}.}\label{alg:two}
\SetKwInOut{KwIn}{Input}
\SetKwInOut{KwOut}{Output}

\tcp{$\triangleright$ Cross-Modal Attribute Insertions}
\KwIn{An image-text pair denoted by $(\mathcal{I}, \mathcal{T})$}
\KwOut{Augmented text $\mathcal{T'}$.}
\tcp{$\triangleright$ Object and Attribute Detection}
Detect objects and attributes in $\mathcal{I}$\\
Introduce masks into $\mathcal{T}$ where direct matches exist\\
If no direct matches, use word similarity b/w detected objects in $\mathcal{I}$ \& nouns in $\mathcal{T}$\\
\tcp{$\triangleright$ Mask Prediction}
\For{$i=1$, ..., $N_{[MASK]}$}{
    Use BERT to obtain top-$k$ predictions for current mask\\
    For current mask, maintain probability score vector $p$\\
    \tcp{$\triangleright$ Attribute Similarity}
    \For{$j=1$, ..., $k$}{
        Compute maximum attribute similarity between relevant object attributes and the current predicted word, $s_j$\\
    }
    \tcp{$\triangleright$ Cross-Modal Dissimilarity for Estimating Robustness}
    Create $k$ candidate augmentations\\
    Compute CLIP dissimilarity for each candidate augmentation, $d_1$, ..., $d_k$\\
    \tcp{$\triangleright$ Text Augmentation Strategy}
    Compute the final score vector, $\mathcal{S}_w$\\
    Insert word with maximum score in $\mathcal{S}_w$ in place of current \texttt{[MASK]}
}
Output text $\mathcal{T'}$ with insertions
\end{algorithm}

\end{document}